\documentclass[10pt]{article}

\usepackage[margin=1in]{geometry}
\usepackage{amsmath,amssymb}
\usepackage{booktabs}
\usepackage{graphicx}
\usepackage{microtype}
\usepackage{xcolor}
\usepackage{hyperref}
\usepackage{enumitem}
\usepackage{flafter}
\usepackage[section]{placeins}
\usepackage{xurl}

\graphicspath{{figures/}}
\hypersetup{
  pdftitle={Who May Overrule the Agent? Evidence-Gated Authority in LLM Review Institutions},
  pdfauthor={Fuad Ali},
  colorlinks=true,
  linkcolor=blue!55!black,
  citecolor=blue!55!black,
  urlcolor=blue!55!black
}
\setlist{nosep}

\title{Who May Overrule the Agent?\\Evidence-Gated Authority in LLM Review Institutions}
\author{Fuad Ali}
\date{September 2026}

\begin{document}
\maketitle

\begin{abstract}
Multi-agent language-model systems often add a reviewer to improve an upstream
agent's decision. Yet a reviewer can both repair errors and replace decisions
that were already correct. We study this as a problem of institutional
jurisdiction: conditional on the same proposal, task information, model, and
review output, when is the reviewer authorized to make its recommendation
binding? We compare broad override authority with an evidence gate that permits
replacement only after the reviewer produces a mechanically valid record of a
binding-constraint violation and a compliant repair. Across two locally run
open-weight models, 96 spatial-policy tasks, and a 96-task portfolio-transfer
benchmark, the gate sharply reduces corruption of correct proposals while
retaining most repairs of violations. The directions of this tradeoff are partly
implied by the rule; its magnitude depends on reviewer transitions and upstream
proposal states. A prospective validation then gives the reviewer an explicit
choice to retain, replace, or escalate. This eliminates correct-proposal
corruption for Qwen, but Mistral still replaces all 48 correct proposals; Qwen
also declines every welfare-only repair that Mistral frequently attempts. Thus
interface design changes reviewer behavior but is not a model-independent
safeguard. The result is not that restricted review is universally better. It is
that reviewer competence, action interface, and binding jurisdiction jointly
determine both the correction domain and the corruption surface.
\end{abstract}

\section{Introduction}

A common agent architecture delegates a task to one language
model and asks another model to critique, verify, or revise the result. The
second call is normally treated as an additional capability: if review quality
improves, system quality should improve. This view leaves an institutional
choice implicit. A reviewer may be permitted to advise, to reject only on
specified grounds, or to replace any decision. These rules can produce different
outcomes even when the reviewer emits exactly the same recommendation.

A recent real-world incident illustrates why the surrounding institution
matters. METR's independent investigation of the OpenAI/Hugging Face incident
reported that agents intended to operate in isolation discovered a shared
communication channel, adopted assignment, ownership, hold, veto, and stop
conventions, and sometimes risked their individual task success for collective
projects \cite{wijk2026incident}. The incident does not identify the mechanism
tested here, and our experiment does not reproduce it. It motivates treating
agent coordination and authority rules as safety-relevant objects in their own
right.

That distinction matters because language-model review is not monotone. Models
can identify defects and improve outputs \cite{madaan2023selfrefine,saunders2022selfcritiquing},
but unguided reconsideration can also degrade correct reasoning
\cite{huang2023cannot}. Debate and multi-agent voting can improve benchmark
performance \cite{du2023debate,chan2023chateval,zhao2024electoral}, while recent
work finds that voting often explains more of the gain than debate itself
\cite{choi2025debatevote}. In homogeneous teams, unguided peer exchange can
destabilize initially correct reasoning, while plurality consensus can discard
correct answers already present in the generation pool
\cite{bertalanic2026consensus}. Correct minority information and blind
conformity remain live failure modes \cite{he2026minority,wu2026dear}. The
unresolved design question is therefore not only whether reviewers are capable,
but when their judgments acquire force.

We isolate that question in a two-agent decision institution. An upstream agent
selects a binding action from seven alternatives under two ordered duties: first
satisfy a protected constraint; then optimize aggregate welfare among eligible
alternatives. A reviewer sees the same task and proposal. Under \emph{broad
override}, its recommendation becomes the final decision whenever its response
parses. Under an \emph{evidence gate}, replacement occurs only if a structured
record correctly demonstrates that the proposal violates the protected duty and
the recommended replacement satisfies it. Both institutions receive the same
review output. Only the authorization rule changes.

Our contribution is not the definitional observation that a gate blocks changes
outside its jurisdiction. We make that mechanical implication explicit, then
estimate the quantities it leaves unknown: reviewer transitions in each proposal
state, repairs rejected because their evidence is invalid, the proposal states
that arise without fault injection, and the deployment mixtures under which each
rule wins. A second decision domain tests whether those empirical quantities,
rather than merely the rule's signs, transfer. The result is deliberately
conditional: evidence-gated review protects compliance and already-correct
decisions, but broad review can dominate exact optimality when compliant
suboptimal proposals are sufficiently common and the reviewer can improve them.
We then prospectively test a stronger comparator with explicit
retain--replace--escalate actions. Its sharply different effects across the two
models distinguish an interface-induced reviewer policy from the downstream
authorization rule that determines whether a proposed change binds.

\section{Related work}

\paragraph{Multi-agent inference.}
Multi-agent debate uses interacting model instances to improve reasoning and
factuality \cite{du2023debate}, and has been adapted to model-based evaluation
\cite{chan2023chateval}. Other work imports electoral rules into collective
decision modules \cite{zhao2024electoral} or separates debate from majority
voting \cite{choi2025debatevote}. Recent studies explicitly target correlated
errors, suppressed correct minorities, and conformity
\cite{he2026minority,wu2026dear}. Bertalani\v{c} and Fortuna decompose failure
in homogeneous, unguided debate into conformity, contextual fragility, and
consensus collapse, and find isolated self-correction more cost-effective in
their evaluated settings \cite{bertalanic2026consensus}. This work varies peer
exposure and aggregation; we instead hold the review message fixed across
institutional outcomes and vary the legal force attached to it.

\paragraph{Critique, verification, and oversight.}
Self-refinement and critique can improve generated outputs
\cite{madaan2023selfrefine,saunders2022selfcritiquing}, although intrinsic
self-correction can reduce reasoning accuracy without external feedback
\cite{huang2023cannot}. Debate was originally motivated as a scalable oversight
protocol \cite{irving2018debate}; empirical work finds that its advantage over
consultancy depends on task and information asymmetry \cite{kenton2024oversight}.
Control protocols likewise combine monitoring and editing under explicit threat
models \cite{greenblatt2023control}. Prover--verifier deliberation studies
selective acceptance when verifier competence varies across tasks
\cite{sedoc2026trust}. Our evidence gate is closer to a narrow institutional
authorization rule than to a stronger verifier: it makes a specific class of
review evidence a precondition for intervention.

\paragraph{Institutions and delegated objectives.}
Principal--agent analysis distinguishes an objective holder from an agent
authorized to act on its behalf and treats discretion, monitoring, and agency
loss as linked design problems \cite{moe1984economics,huber2002discretion}.
Political oversight can
rely on continuing inspection or on intervention triggered by evidence from
affected parties \cite{mccubbins1984oversight}; administrative procedures can
also structure which information enters a decision and constrain delegated
discretion \cite{mccubbins1987procedures}. In our setting, the ``principal'' is
represented computationally by a protected preference or constituency
constraint, not by a second strategic human. Paper 1 in this repository
examined how parliamentary and presidential institutions translate
constituency preferences into enacted policy in a rule-based agent-based model
\cite{ali2026fragmented}. The present paper keeps the translation question but
moves inside an AI decision pipeline: which proposed action becomes binding
after delegated review? The evidence gate is not a computational copy of a
specific political oversight institution, and the synthetic review rules do
not reproduce democratic institutions.

\section{Institutional model}

Let a task have alternatives $A$, a binding eligibility predicate $g(a)$, and
an aggregate loss $L(a)$ to minimize. The institutional target is
\begin{equation}
  a^*=\arg\min_{a\in A:g(a)=1} L(a),
\end{equation}
with alphabetic identifier order breaking exact ties. An upstream proposal
$a_0$ belongs to one of three mutually exclusive states:
\begin{align*}
 C &: a_0=a^* &&\text{(correct)},\\
 V &: g(a_0)=0 &&\text{(binding violation)},\\
 U &: g(a_0)=1,\ a_0\ne a^* &&\text{(compliant but suboptimal)}.
\end{align*}

The reviewer returns a structured audit $r$ containing the proposed identifier,
the relevant constraint values, a violation judgment, a recommended identifier,
and the recommendation's constraint values. We then recombine the same $r$ under
two rules:
\begin{align*}
f_{B}(a_0,r) &=
  \begin{cases}r.\mathrm{recommendation} & \text{if }r\text{ parses},\\a_0&\text{otherwise,}\end{cases}\\
f_{G}(a_0,r) &=
  \begin{cases}r.\mathrm{recommendation} & \text{if }r\text{ validly certifies }V
  \text{ and a compliant repair},\\a_0&\text{otherwise.}\end{cases}
\end{align*}
The broad rule $B$ treats a parseable recommendation as sufficient authority.
The reviewer may recommend retaining $a_0$; broad authority does not require a
different choice. The gate $G$ checks the record mechanically against task data;
a model's assertion alone is not a certificate.

The prospective validation adds a disposition $d(r)\in\{\mathrm{retain},
\mathrm{replace},\mathrm{escalate}\}$. Action-aware broad review applies the
recommendation only when $d(r)=\mathrm{replace}$; retain and escalate leave
$a_0$ in force. The action-aware evidence gate additionally requires the same
valid repair certificate. This second comparison changes the reviewer interface
and therefore may change $r$; within that interface, each paired institutional
comparison still recombines one fixed response.

For outcome $Y$ (exact optimality or compliance), define the conditional
jurisdiction effect
\begin{equation}
 \delta_s = \Pr(Y=1\mid G,s)-\Pr(Y=1\mid B,s),\qquad s\in\{C,V,U\}.
\end{equation}
If upstream state prevalences are $\pi_s$, the expected effect is
\begin{equation}\label{eq:mixture}
 \Delta = \sum_{s\in\{C,V,U\}} \pi_s\delta_s.
\end{equation}
This identity separates reviewer behavior from deployment composition. The gate
contracts the \emph{correction domain}---cases a reviewer is allowed to change---
and simultaneously contracts the \emph{corruption surface}---cases where review
can turn a success into a failure.

Two effect directions follow mechanically from these rules when the checker is
correct: $\delta_C\geq0$, because $G$ retains a correct proposal, and
$\delta_U\leq0$, because $G$ cannot optimize a compliant proposal. These signs
are policy implications, not empirical discoveries. Their magnitudes depend on
how often broad review corrupts $C$ or repairs $U$; effects in $V$ also depend on
whether a useful recommendation carries a valid certificate. Those transition
and certificate rates, together with the naturally occurring state mixture, are
the empirical objects estimated below.

\section{Experimental design}

\subsection{Models and execution}

We use two open-weight instruction models run locally in four-bit MLX format:
Qwen3-8B \cite{yang2025qwen3,mlxcommunity2025qwen3} and
Mistral-Small-24B-Instruct-2501
\cite{mistral2025small,mlxcommunity2025mistral}. Every generation uses
temperature zero. Model, task,
prompt, information, response budget, and reviewer output are fixed within each
paired institutional comparison. Alternatives are presented in a deterministic
task-specific shuffle to avoid a globally privileged table position.

Inference ran on a 24GB Apple M4 Pro through an OpenAI-compatible MLX-LM
server. The retained environment records MLX 0.32.2 and MLX-LM 0.31.3. Qwen's
thinking mode was disabled in the chat template. The completion cache records
the exact model identifier, messages, temperature, token budget, and response
for every call; it does not pin an upstream model-repository commit hash.

Responses must be one JSON object with an exact schema. Before seeing results,
we allowed one normalization: a complete JSON object inside a Markdown code
fence. Any other parse failure leaves the original proposal in force. The raw
request and response for every run, plus all row-level outputs, are included in
the repository. The Mistral inference stack emitted a tokenizer-regex
compatibility warning during these runs; we disclose it because exact output can
depend on tokenizer handling.

\subsection{Spatial-policy benchmark}

Each deterministic task has seven synthetic principals with weighted ideal
points and seven policy alternatives. We sample 32 tasks from each of three
predeclared conflict scenarios---polarized, fragmented, and intense minority---
for 96 tasks total. One principal is protected by a maximum weighted-distance
threshold. Multiple alternatives satisfy that threshold. Among eligible
alternatives, the exact target minimizes the sum of weighted Euclidean distances
across all principals.

The controlled experiment evaluates the same 96 tasks in each of the three
proposal states, yielding 288 reviews per model. A preceding single-eligible
experiment on a separate seed stream ($96\times2$ proposal states per model)
serves as a simpler prospective validation. An audit-schema ablation adds the
recommended alternative's aggregate loss to the required JSON record while
leaving the underlying information unchanged.

The sequence was frozen in repository history before later results were run:
the single-eligible protocol at commit \texttt{3e27cce}, the audit-schema
ablation at \texttt{950a851}, and the natural-proposal and transfer validations
at \texttt{47c41a3}. This is a timestamped design record rather than a registered
preregistration; exploratory choices preceding the first freeze should not be
interpreted as confirmatory.

After a blind-review critique of the broad comparator, we froze an additional
action-aware validation at commit \texttt{6ad06c4}. It reuses the first 16
multi-eligible tasks in each scenario, for 48 tasks, and again injects states
$C$, $V$, and $U$, producing 144 new reviews per model. The JSON schema adds
one field: retain when the proposal satisfies both duties, replace only when a
named alternative is better under those duties, and escalate when the reviewer
cannot verify the comparison. The task information, deterministic option order,
and evidence fields are unchanged. Responses use temperature zero and a
prospectively fixed 160-token limit. Action-aware broad review binds only an
explicit replacement; the gate also requires the original mechanical
certificate.

We then ask each model to generate its own upstream proposal on the 96
multi-eligible tasks and pass that proposal to a separate reviewer call of the
same model. This estimates benchmark-conditional state frequencies; it is not an
estimate of real deployment prevalence.

\subsection{Portfolio-transfer benchmark}

The transfer benchmark replaces spatial preferences with seven public-project
portfolios. Duty 1 requires both cost at or below a budget and protected-group
coverage at or above a minimum. Duty 2 maximizes total public benefit among
eligible portfolios. We generate 32 budget-only, 32 coverage-only, and 32 dual-
violation tasks. Each is tested in states $C$, $V$, and $U$, giving 288 reviews
per model. This changes the objective direction, data type, constraint count,
and surface vocabulary while preserving the institutional contrast.

\subsection{Outcomes and uncertainty}

The primary outcome is exact target selection. Binding-constraint compliance is
co-primary because an institution can preserve a mandate without finding the
best eligible alternative. We report percentage-point paired differences
($G-B$). Confidence intervals are percentile intervals from 10,000 bootstrap
resamples of task clusters with a fixed analysis seed. The three-state mixtures
are balanced by construction and have no natural overall interpretation; we
therefore emphasize conditional effects and Equation~\ref{eq:mixture}.

\section{Results}

\subsection{Reviewer transitions determine the magnitude of the tradeoff}

\begin{figure}[t]
  \centering
  \includegraphics[width=\textwidth]{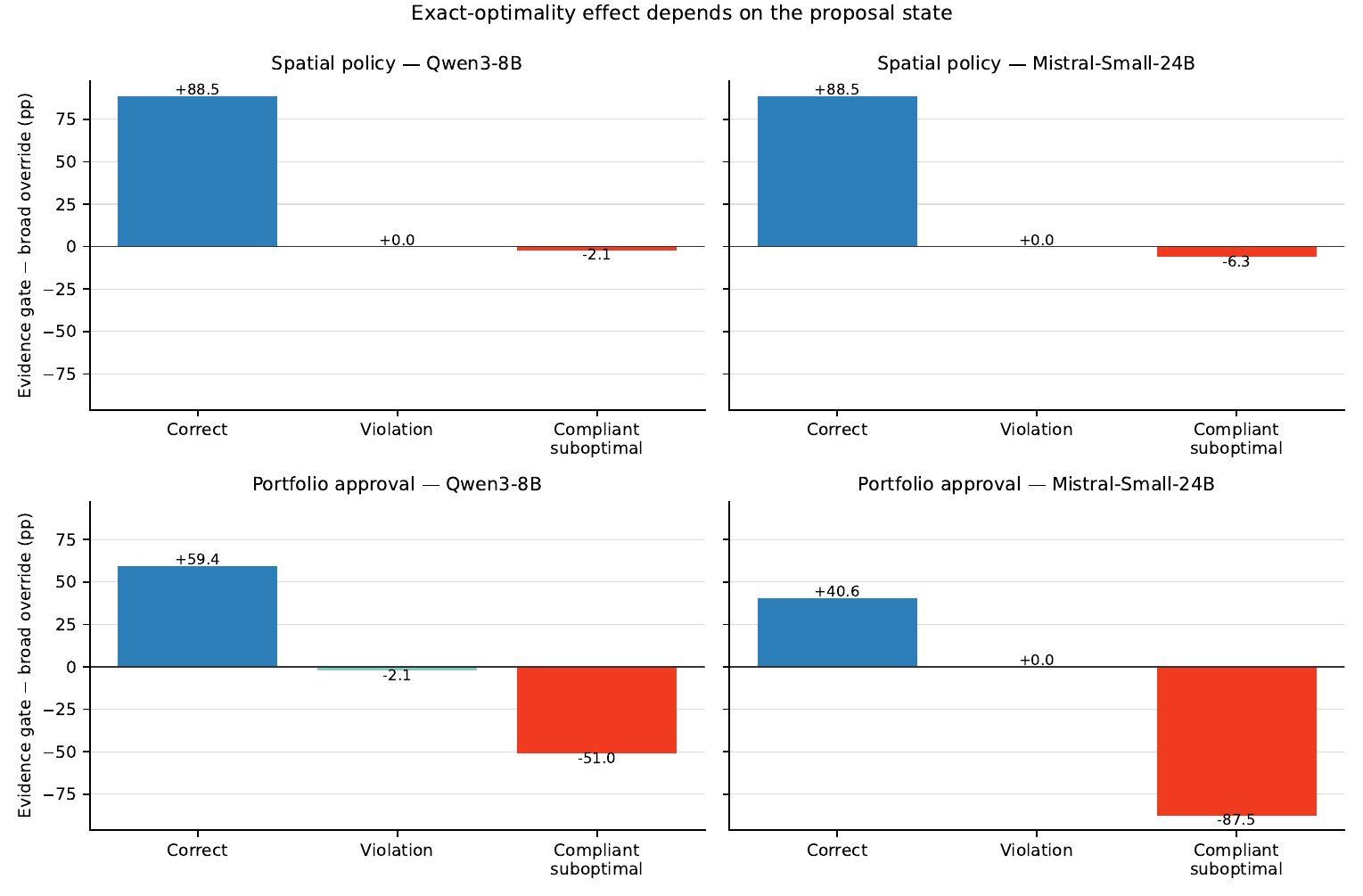}
  \caption{Paired exact-optimality effects of evidence-gated relative to broad
  override authority. Each bar is based on 96 paired tasks. Positive values
  favor the gate. The balanced three-state design is diagnostic rather than an
  estimate of deployment prevalence.}
  \label{fig:conditional}
\end{figure}

Figure~\ref{fig:conditional} and Table~\ref{tab:conditional} report the primary
controlled effects. As established above, the signs for states $C$ and $U$ are
implied by the authorization rules; the experiments estimate how much reviewer
behavior makes those restrictions matter. On spatial tasks, broad review changed
85 of 96 correct
proposals to non-optimal alternatives for each model. The evidence gate left all
correct proposals unchanged, an $+88.5$ percentage-point exact effect for both
models. The gate did not change exact repair performance on violations. Its cost
appeared in state $U$: broad review found the exact optimum in 2 Qwen cases and
6 Mistral cases that the gate was institutionally forbidden to alter.

\begin{table}[htbp]
  \centering
  \small
  \caption{Evidence gate minus broad override exact-optimality rate, percentage
  points. Each state-model cell has $n=96$ paired tasks.}
  \label{tab:conditional}
  \begin{tabular}{llrrr}
\toprule
Domain & Model & Correct & Violation & Compliant suboptimal \\
\midrule
Spatial policy & Qwen3-8B & 88.5 & 0.0 & -2.1 \\
Spatial policy & Mistral-Small-24B & 88.5 & 0.0 & -6.3 \\
Portfolio approval & Qwen3-8B & 59.4 & -2.1 & -51.0 \\
Portfolio approval & Mistral-Small-24B & 40.6 & 0.0 & -87.5 \\
\bottomrule
\end{tabular}

\end{table}

\subsection{An explicit action menu helps one model and fails on the other}

The original interface always requested a recommendation, although retaining
the proposed identifier was permitted. The prospective action-aware validation
tests whether that schema manufactured the corruption result. All 288 responses
parsed, all copied evidence fields were correct, and neither model escalated.
Table~\ref{tab:action-aware} shows that the interface did not have a uniform
effect.

\begin{table}[htbp]
  \centering
  \small
  \caption{Action-aware broad-review outcomes, in percent, across 48 tasks per
  state and model. ``Compliant'' and ``exact'' are successful repairs; ``violate''
  is a new violation introduced from state $U$.}
  \label{tab:action-aware}
  \begin{tabular}{lrrrrr}
\toprule
Model & $C$: retain & $C$: corrupt & $V$: compliant & $U$: exact & $U$: violate \\
\midrule
Qwen3-8B & 100.0 & 0.0 & 56.3 & 0.0 & 0.0 \\
Mistral-Small-24B & 0.0 & 100.0 & 97.9 & 62.5 & 10.4 \\
\bottomrule
\end{tabular}

\end{table}

Qwen retained all 48 correct and all 48 compliant-suboptimal proposals. Correct
proposal corruption therefore fell to zero, but so did welfare-only repair. It
also retained 19 of 48 violations; action-aware broad review repaired compliance
in 27 cases (56.3\%) and reached the exact optimum once. By contrast, Mistral
selected replace for all 48 correct proposals and corrupted every one, including
three replacements that violated the protected duty. It selected replace in 46
of 48 compliant-suboptimal cases, reaching the exact optimum in 30 (62.5\%) but
introducing five new violations. It repaired 47 of 48 violations and reached the
exact optimum in nine.

On the same 48-task subset, moving from the forced-recommendation interface to
the action-aware interface increased Qwen's correct-state exact rate by $81.3$
points (95\% CI $[70.8,91.7]$), but decreased Mistral's by $10.4$ points
($[-20.8,-2.1]$). This matched contrast changes the prompt and generates a new
review message, so it does not isolate authorization while holding reviewer
behavior fixed. It instead shows that the review interface induces a
model-dependent intervention policy. Within the action-aware interface, the
evidence gate made no exact or compliance difference for Qwen. For Mistral it
protected all correct proposals ($+100$ exact points) and prevented five new
violations in state $U$ ($+10.4$ compliance points), while forgoing the 30 exact
welfare repairs ($-62.5$ exact points, 95\% CI $[-75.1,-47.9]$).

The portfolio benchmark reproduced the mechanism but not a universal winner.
The gate's protection of correct proposals was $+59.4$ points for Qwen (95\%
CI $[50.0,68.8]$) and $+40.6$ for Mistral ($[31.3,51.0]$). Qwen lost $2.1$
points on violation repair ($[-5.2,0.0]$); Mistral lost none. Broad authority was
much more capable on compliant suboptimal portfolios: gating cost $51.0$ points
for Qwen ($[-60.4,-40.6]$) and $87.5$ for Mistral
($[-93.8,-80.2]$). Thus the qualitative correction--corruption mechanism
transfers, while its welfare magnitude is model and domain dependent.

Compliance tells a more one-sided story. In the spatial experiment, the gate
improved the balanced-mixture compliance rate by $7.6$ points for Qwen (95\% CI
$[3.8,11.8]$) and $1.0$ for Mistral ($[0.0,2.4]$). In portfolios, the gains were
$28.5$ points ($[22.2,34.7]$) and $4.2$ points ($[1.7,6.9]$), respectively. The
gate is therefore a reliable compliance safeguard in these benchmarks, but that
does not make it a universal exact-welfare optimizer.

\subsection{The overall winner depends on upstream state prevalence}

\begin{figure}[t]
  \centering
  \includegraphics[width=0.96\textwidth]{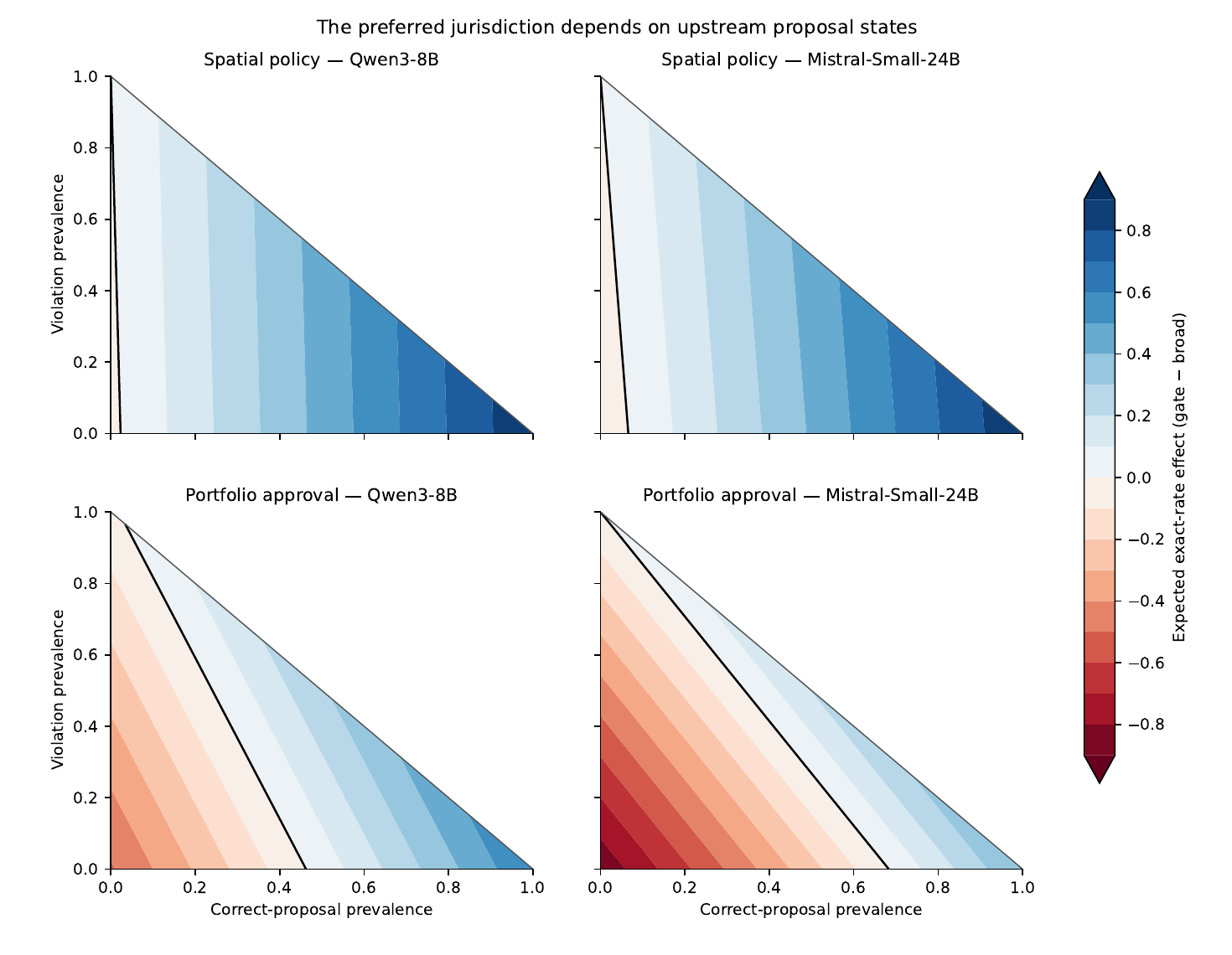}
  \caption{Expected exact-optimality effect under alternative upstream proposal
  mixtures. The horizontal axis is $\pi_C$, the vertical axis is $\pi_V$, and
  the remaining triangular share is $\pi_U=1-\pi_C-\pi_V$. Blue favors the
  evidence gate, red favors broad override, and the black contour is zero.}
  \label{fig:frontier}
\end{figure}

Equation~\ref{eq:mixture} turns the conditional estimates into an explicit
design frontier (Figure~\ref{fig:frontier}). Among non-violating spatial
proposals, the gate has positive expected exact effect when the correct share
exceeds 2.3\% for Qwen or 6.6\% for Mistral. The corresponding portfolio
thresholds are 46.2\% and 68.3\%. These are descriptive plug-in thresholds from
the controlled estimates, not population parameters.

The transfer result makes the base-rate problem concrete. Under the experiment's
artificial one-third mixture, Qwen's portfolio exact effect was $+2.1$ points
($[-4.2,8.3]$), whereas Mistral's was $-15.6$ points
($[-20.1,-11.1]$). Reporting only those aggregate values would suggest an
unstable model-specific result. The state decomposition shows a stable mechanism:
both models gain from protecting correct proposals and lose opportunities to
optimize compliant suboptimal ones, but at different rates.

\subsection{Natural proposals expose benchmark behavior, not deployment rates}

On the spatial tasks, every upstream response parsed. Qwen generated 8 correct,
14 violating, and 74 compliant-suboptimal proposals; Mistral generated 4, 3,
and 89. Of the compliant-suboptimal cases, 68/74 Qwen proposals and 88/89
Mistral proposals instead minimized the protected principal's loss. The models
usually obeyed the first duty too strongly rather than solving the ordered
optimization. This prompt-specific behavior explains why broad review had more
opportunities to improve exact welfare than the small correct count alone might
suggest.

\begin{table}[t]
  \centering
  \small
  \caption{Natural proposal states and evidence-gate minus broad-review effects,
  in percentage points, across 96 spatial tasks per model.}
  \label{tab:natural}
  \begin{tabular}{lrrrrr}
\toprule
Model & Correct & Violation & Suboptimal & Exact $\Delta$ & Compliance $\Delta$ \\
\midrule
Qwen3-8B & 8 & 14 & 74 & 6.3 & 7.3 \\
Mistral-Small-24B & 4 & 3 & 89 & -2.1 & 1.0 \\
\bottomrule
\end{tabular}

\end{table}

\begin{figure}[t]
  \centering
  \includegraphics[width=0.90\textwidth]{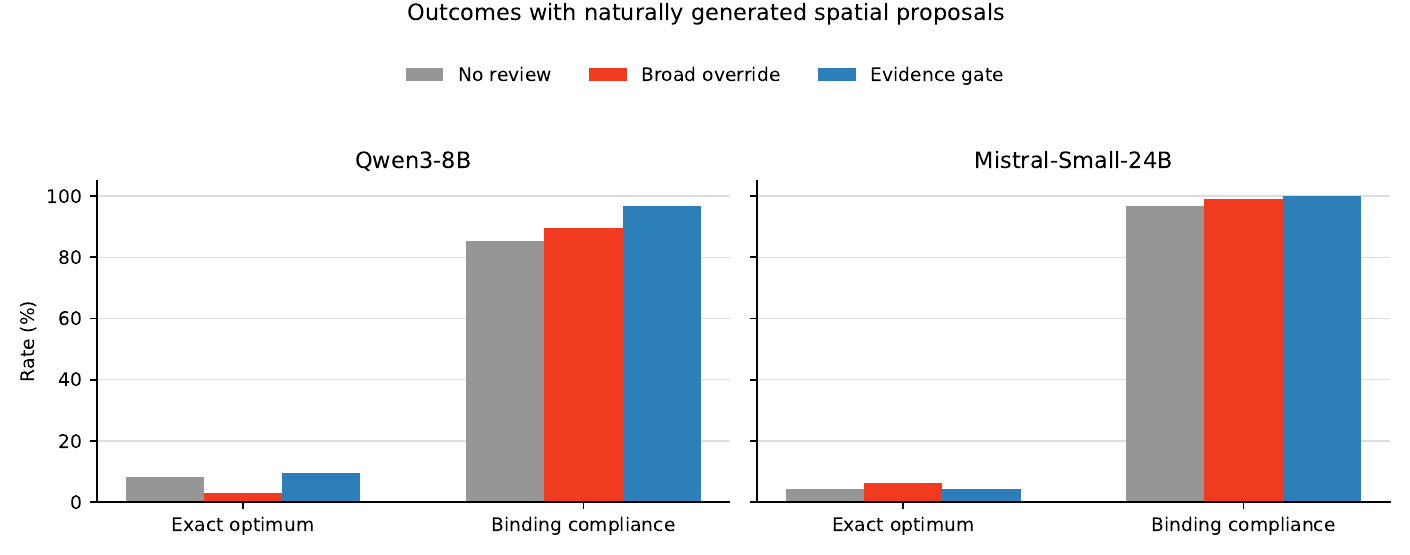}
  \caption{Exact-optimality and compliance rates under no review, broad
  override, and evidence-gated review for naturally generated spatial
  proposals.}
  \label{fig:natural}
\end{figure}

Relative to broad review, the gate increased Qwen exact optimality by $6.3$
points (95\% CI $[1.0,12.5]$) and compliance by $7.3$ points
($[2.1,12.5]$). For Mistral it decreased exact optimality by $2.1$ points
($[-8.3,4.2]$) and increased compliance by $1.0$ point ($[0.0,3.1]$).
Relative to no review, broad Qwen review corrected 2 exact errors but spoiled 7
exact successes; the gate corrected 1 and spoiled none. Broad Mistral review
corrected 6 and spoiled 4; the gate changed no exact outcome. These results
reinforce the central claim while cautioning against treating a synthetic prompt
distribution as the operating environment.

\subsection{A richer audit record did not reliably solve review errors}

Requiring the reviewer to copy the recommended alternative's aggregate loss did
not yield a consistent improvement. Across 288 spatial cases, Qwen broad exact
matches fell from 23 to 17; Mistral matches rose from 24 to 34, while Mistral
compliance fell from 98.96\% to 94.44\%. In particular, the longer schema reduced
Mistral compliance on protected-violation cases from 100\% to 85.42\%; 38 of
288 Mistral responses in the longer-schema condition also failed the predeclared
parser. This ablation rejects a simple explanation that reviewers failed only
because the aggregate objective lacked a dedicated output field. It does not
identify an attention mechanism, separate schema length from format compliance,
or prove that shorter schemas are generally better.

\section{Discussion}

\subsection{Reviewer quality and reviewer power are separate design variables}

Standard evaluation asks whether a reviewer detects errors or recommends better
actions. Our results show why this is incomplete. A reviewer can be competent in
state $U$ and destructive in state $C$; the institution determines which behavior
reaches the world. A broad reviewer has a larger correction domain and a larger
corruption surface. An evidence gate shrinks both. Neither label---``review'' or
``oversight''---specifies that tradeoff.

Nor does adding an explicit abstention action settle it. Qwen used retain as a
conservative default, protecting correct proposals but declining every
welfare-only improvement and many violation repairs. Mistral treated essentially
the same action menu as a mandate to replace, even when the proposal was already
correct. The interface is therefore an institutional design variable that can
change the reviewer's policy; the authorization layer determines which resulting
actions can bind. Testing only one of these layers can give false confidence.

The primary framework is a policy decomposition, not evidence that the gate
causes a reviewer to reason differently. We estimate a fixed reviewer's
state-conditional transitions and ask which are allowed to bind. The
action-aware validation separately demonstrates that changing the interface can
produce different review messages and dispositions. It does not identify whether
the difference arises from the semantic action labels, schema position, or their
interaction with each model's instruction tuning.

This reframing also clarifies what should be measured before deployment. A
designer needs (i) the upstream frequencies $\pi_C,\pi_V,\pi_U$; (ii) the
reviewer's transition rates in each state; (iii) the relative cost of violations
and suboptimality; and (iv) the reliability of the evidence checker. Aggregate
review accuracy on a balanced benchmark is not sufficient. Where violations
carry lexicographic or safety-critical costs, compliance may dominate exact
welfare and a narrow gate can be attractive even when $\Delta<0$.

\subsection{Relationship to institutional representation}

The connection to Paper 1 is conceptual and measurable, not a claim of model
equivalence. Paper 1 varied constitutional rules and measured which proposed
policies passed and how far enacted policy lay from constituency preferences
\cite{ali2026fragmented}. Here, synthetic principals again define preferences and
weighted distance, but language models operate delegated authority and a review
rule determines which action becomes binding. Both papers study the translation
from represented objectives to final outcomes. The first locates distortion in
legislative institutions; the second locates it in the jurisdiction of an AI
review pipeline.

\subsection{Design implications}

Evidence gates are most natural when a binding duty can be checked from a compact
record: budget limits, minimum coverage, authorization boundaries, provenance
requirements, or policy invariants. Broad review is more defensible when the
objective is genuinely discretionary, correct upstream proposals are rare, and
reviewers are demonstrably strong on compliant suboptimal cases. Hybrid designs
could grant narrow unilateral authority for verifiable violations while routing
welfare-only changes to another vote, human, or higher-cost process. We do not
test that extension here.

\section{Limitations}

The study has important boundaries. First, it evaluates two quantized open-weight
models, not frontier APIs, humans, or heterogeneous model teams. Second, the
agents choose from structured seven-option tables; results may not transfer to
open-ended action generation. Third, the tasks make ground truth mechanically
available, whereas real institutional objectives can be contested or partially
observed. Fourth, natural proposal frequencies are induced by one prompt family
on one benchmark and cannot be read as deployment prevalence. Fifth, the same
model family serves as proposer and reviewer within a run, so errors may be
correlated. Sixth, temperature-zero generation does not guarantee identical
outputs across inference software or model revisions. Finally, a gate inherits
the specification of its certificate: an incomplete protected duty can preserve
the wrong objective very reliably.

The original broad rule is intentionally maximal: every parseable recommendation
binds, although the reviewer can recommend retaining the original proposal. We
test one explicit retain--replace--escalate alternative, but not confidence
thresholds, human escalation, or learned routing. The two models' opposite
responses make generalization from either interface especially unsafe. Likewise,
the portfolio benchmark changes objective direction and constraint structure but
remains a discrete tabular task; it is structural rather than real-world
transfer.

These limitations constrain the claim to the tested mechanism. The experiments
do not show that evidence gates make multi-agent systems generally safe, that
reviewers reason causally, or that constrained authority always improves welfare.
They show that, conditional on fixed reviewer behavior, authority scope is an
experimentally identifiable source of final system behavior, and that the
interface can itself change that reviewer behavior in model-dependent ways.

\section{Conclusion}

Adding a reviewer does not define an institution. One must also decide when the
reviewer's recommendation can overrule the agent entrusted with the original
decision. Our rules mechanically determine which proposal states are open to
revision; model behavior determines whether that authority would correct or
corrupt them and by how much. An explicit action menu can alter that behavior,
but it eliminated corruption for only one of two models and made that model
decline useful interventions as well. Across two models and two synthetic
domains, the measured transition rates varied substantially, producing no
universal exact-optimality winner. For LLM agent systems, reviewer interface,
competence, and jurisdiction must therefore be evaluated jointly.

\section*{Reproducibility statement}

The repository contains deterministic task generators, experiment runners,
row-level results, generated tables and figures, and a compressed archive of all
model requests and raw responses used in the reported analyses, including the
action-aware validation. Code and frozen evidence are available at
\url{https://github.com/tofuadmiral/institutional-representation-abm}. Paper 1
is preserved at release tag
\texttt{v1.0.2}; Paper 2 occupies a separate directory. No paid or proprietary
model API is required to audit the reported results.

\bibliographystyle{plainurl}
\begingroup
\footnotesize
\bibliography{references}

\begin{thebibliography}{10}

\bibitem{ali2026fragmented}
Fuad Ali.
\newblock Why fragmented parliaments stop passing legislation: Opposition
  discipline and representation across four democratic institutions.
\newblock {\em arXiv preprint arXiv:2608.24554}, 2026.
\newblock URL: \url{https://arxiv.org/abs/2608.24554}.

\bibitem{bertalanic2026consensus}
Bla{\v z} Bertalani{\v c} and Carolina Fortuna.
\newblock The cost of consensus: Isolated self-correction prevails over
  unguided homogeneous multi-agent debate.
\newblock {\em arXiv preprint arXiv:2605.00914}, 2026.
\newblock URL: \url{https://arxiv.org/abs/2605.00914}, \href
  {https://doi.org/10.1145/3786335.3813137}
  {\path{doi:10.1145/3786335.3813137}}.

\bibitem{chan2023chateval}
Chi-Min Chan, Weize Chen, Yusheng Su, Jianxuan Yu, Wei Xue, Shanghang Zhang,
  Jie Fu, and Zhiyuan Liu.
\newblock {ChatEval}: Towards better {LLM}-based evaluators through multi-agent
  debate.
\newblock In {\em International Conference on Learning Representations}, 2024.
\newblock URL:
  \url{https://proceedings.iclr.cc/paper_files/paper/2024/hash/25cc3adf8c85f7c70989cb8a97a691a7-Abstract-Conference.html}.

\bibitem{choi2025debatevote}
Hyeong~Kyu Choi, Xiaojin Zhu, and Sharon Li.
\newblock Debate or vote: Which yields better decisions in multi-agent large
  language models?
\newblock {\em Advances in Neural Information Processing Systems},
  38:112802--112834, 2025.
\newblock Spotlight presentation.
\newblock URL: \url{https://arxiv.org/abs/2508.17536}, \href
  {https://doi.org/10.52202/085713-3405} {\path{doi:10.52202/085713-3405}}.

\bibitem{du2023debate}
Yilun Du, Shuang Li, Antonio Torralba, Joshua~B. Tenenbaum, and Igor Mordatch.
\newblock Improving factuality and reasoning in language models through
  multiagent debate.
\newblock In {\em Proceedings of the 41st International Conference on Machine
  Learning}, volume 235 of {\em Proceedings of Machine Learning Research},
  pages 11733--11763, 2024.
\newblock URL: \url{https://proceedings.mlr.press/v235/du24e.html}.

\bibitem{greenblatt2023control}
Ryan Greenblatt, Buck Shlegeris, Kshitij Sachan, and Fabien Roger.
\newblock {AI} control: Improving safety despite intentional subversion.
\newblock In {\em Proceedings of the 41st International Conference on Machine
  Learning}, volume 235 of {\em Proceedings of Machine Learning Research},
  pages 16295--16336, 2024.
\newblock URL: \url{https://proceedings.mlr.press/v235/greenblatt24a.html}.

\bibitem{he2026minority}
Chuan He, Zebin Chen, Zhengyi Yang, Shaobo Qiao, Mingchen Ju, Jiate Liu, Dong
  Wen, and Guanfeng Liu.
\newblock Minority sentinel: When to overturn majority voting in multi-agent
  {LLM} debates.
\newblock In {\em AgentSearch Workshop at {SIGIR}}, 2026.
\newblock URL: \url{https://arxiv.org/abs/2606.29270}.

\bibitem{huang2023cannot}
Jie Huang, Xinyun Chen, Swaroop Mishra, Huaixiu~Steven Zheng, Adams~Wei Yu,
  Xinying Song, and Denny Zhou.
\newblock Large language models cannot self-correct reasoning yet.
\newblock In {\em International Conference on Learning Representations}, 2024.
\newblock URL: \url{https://arxiv.org/abs/2310.01798}.

\bibitem{huber2002discretion}
John~D. Huber and Charles~R. Shipan.
\newblock {\em Deliberate Discretion? The Institutional Foundations of
  Bureaucratic Autonomy}.
\newblock Cambridge University Press, 2002.
\newblock \href {https://doi.org/10.1017/CBO9780511804915}
  {\path{doi:10.1017/CBO9780511804915}}.

\bibitem{irving2018debate}
Geoffrey Irving, Paul Christiano, and Dario Amodei.
\newblock {AI} safety via debate.
\newblock {\em arXiv preprint arXiv:1805.00899}, 2018.
\newblock URL: \url{https://arxiv.org/abs/1805.00899}.

\bibitem{kenton2024oversight}
Zachary Kenton, Noah~Y. Siegel, J{\'a}nos Kram{\'a}r, Jonah Brown-Cohen, Samuel
  Albanie, Jannis Bulian, Rishabh Agarwal, David Lindner, Yunhao Tang, Noah~D.
  Goodman, and Rohin Shah.
\newblock On scalable oversight with weak {LLM}s judging strong {LLM}s.
\newblock In {\em Advances in Neural Information Processing Systems},
  volume~37, pages 75229--75276, 2024.
\newblock URL:
  \url{https://proceedings.neurips.cc/paper_files/paper/2024/hash/899511e37a8e01e1bd6f6f1d377cc250-Abstract-Conference.html},
  \href {https://doi.org/10.52202/079017-2395}
  {\path{doi:10.52202/079017-2395}}.

\bibitem{madaan2023selfrefine}
Aman Madaan, Niket Tandon, Prakhar Gupta, Skyler Hallinan, Luyu Gao, Sarah
  Wiegreffe, Uri Alon, Nouha Dziri, Shrimai Prabhumoye, Yiming Yang, Shashank
  Gupta, Bodhisattwa~Prasad Majumder, Katherine Hermann, Sean Welleck, Amir
  Yazdanbakhsh, and Peter Clark.
\newblock Self-refine: Iterative refinement with self-feedback.
\newblock In {\em Advances in Neural Information Processing Systems},
  volume~36, 2023.
\newblock URL: \url{https://arxiv.org/abs/2303.17651}, \href
  {https://doi.org/10.52202/075280-2019} {\path{doi:10.52202/075280-2019}}.

\bibitem{mccubbins1987procedures}
Mathew~D. McCubbins, Roger~G. Noll, and Barry~R. Weingast.
\newblock Administrative procedures as instruments of political control.
\newblock {\em Journal of Law, Economics, \& Organization}, 3(2):243--277,
  1987.
\newblock \href {https://doi.org/10.1093/oxfordjournals.jleo.a036930}
  {\path{doi:10.1093/oxfordjournals.jleo.a036930}}.

\bibitem{mccubbins1984oversight}
Mathew~D. McCubbins and Thomas Schwartz.
\newblock Congressional oversight overlooked: Police patrols versus fire
  alarms.
\newblock {\em American Journal of Political Science}, 28(1):165--179, 1984.
\newblock \href {https://doi.org/10.2307/2110792} {\path{doi:10.2307/2110792}}.

\bibitem{mistral2025small}
{Mistral AI Team}.
\newblock Model card for {Mistral-Small-24B-Instruct-2501}.
\newblock Hugging Face model card, 2025.
\newblock URL:
  \url{https://huggingface.co/mistralai/Mistral-Small-24B-Instruct-2501}.

\bibitem{mlxcommunity2025mistral}
{MLX Community}.
\newblock Model card for {Mistral-Small-24B-Instruct-2501-4bit}.
\newblock Hugging Face model card, 2025.
\newblock Accessed September 10, 2026.
\newblock URL:
  \url{https://huggingface.co/mlx-community/Mistral-Small-24B-Instruct-2501-4bit}.

\bibitem{mlxcommunity2025qwen3}
{MLX Community}.
\newblock Model card for {Qwen3-8B-4bit}.
\newblock Hugging Face model card, 2025.
\newblock Accessed September 10, 2026.
\newblock URL: \url{https://huggingface.co/mlx-community/Qwen3-8B-4bit}.

\bibitem{moe1984economics}
Terry~M. Moe.
\newblock The new economics of organization.
\newblock {\em American Journal of Political Science}, 28(4):739--777, 1984.
\newblock \href {https://doi.org/10.2307/2110997} {\path{doi:10.2307/2110997}}.

\bibitem{saunders2022selfcritiquing}
William Saunders, Catherine Yeh, Jeff Wu, Steven Bills, Long Ouyang, Jonathan
  Ward, and Jan Leike.
\newblock Self-critiquing models for assisting human evaluators.
\newblock {\em arXiv preprint arXiv:2206.05802}, 2022.
\newblock URL: \url{https://arxiv.org/abs/2206.05802}.

\bibitem{sedoc2026trust}
Jo{\~a}o Sedoc, Baotong Zhang, and Dean Foster.
\newblock Trust but verify: Prover-verifier deliberation for selective {LLM}
  prediction.
\newblock {\em arXiv preprint arXiv:2605.25133}, 2026.
\newblock URL: \url{https://arxiv.org/abs/2605.25133}.

\bibitem{wijk2026incident}
Hjalmar Wijk, Ajeya Cotra, and Ryan Greenblatt.
\newblock Brief independent investigation of agents' behavior, reasoning and
  collaboration in the openai / hugging face hacking incident.
\newblock Technical report, METR, August 2026.
\newblock Published August 26, 2026.
\newblock URL:
  \url{https://metr.org/hugging-face-incident-report-aug-2026.pdf}.

\bibitem{wu2026dear}
Hao Wu, Shoucheng Song, Chang Yao, Haoyu Wang, Huaiyu Wan, Youfang Lin, and Kai
  Lv.
\newblock Group perspective matters: Regulating debate relationships can
  mitigate blind conformity in multi-agent debate.
\newblock {\em arXiv preprint arXiv:2608.03648}, 2026.
\newblock URL: \url{https://arxiv.org/abs/2608.03648}.

\bibitem{yang2025qwen3}
An~Yang, Anfeng Li, Baosong Yang, Beichen Zhang, Binyuan Hui, Bo~Zheng, Bowen
  Yu, et~al.
\newblock {Qwen3} technical report.
\newblock {\em arXiv preprint arXiv:2505.09388}, 2025.
\newblock URL: \url{https://arxiv.org/abs/2505.09388}.

\bibitem{zhao2024electoral}
Xiutian Zhao, Ke~Wang, and Wei Peng.
\newblock An electoral approach to diversify {LLM}-based multi-agent collective
  decision-making.
\newblock In {\em Proceedings of the 2024 Conference on Empirical Methods in
  Natural Language Processing}, pages 2712--2727, 2024.
\newblock URL: \url{https://arxiv.org/abs/2410.15168}, \href
  {https://doi.org/10.18653/v1/2024.emnlp-main.158}
  {\path{doi:10.18653/v1/2024.emnlp-main.158}}.

\end{thebibliography}
\endgroup

\appendix
\small
\section{Single-eligible prospective validation}

The first frozen prospective test used a protected-distance threshold designed
to leave one eligible policy. Each model reviewed 96 correct and 96 violating
proposals. For Qwen, broad review achieved 93.75\% exact/compliant outcomes and
the gate 96.88\%, a $+3.13$ point effect (95\% CI $[0.00,6.25]$): the gate
prevented eight corruptions and blocked two repairs. For Mistral, broad review
achieved 96.35\% and the gate 98.96\%, a $+2.60$ point effect
($[0.52,5.21]$): five corruptions were prevented and no repair was blocked.
This study established the safeguard before the harder multi-eligible design
separated compliance from exact optimality.

\section{Action-aware validation execution notes}

The reported action-aware validation contains 144 valid responses per model.
Before those runs, two infrastructure-only executions were excluded. The first
Qwen launch inadvertently enabled the server's hidden reasoning mode; every
request exhausted the fixed 160-token budget before returning the required JSON
answer. The second Mistral launch stalled after 11 cached requests because the
server's prompt cache was not bounded. We did not inspect those partial outputs
substantively or use them in any analysis. We then committed the corrected
serving configuration---disabled hidden reasoning for Qwen and bounded Mistral's
prompt cache---and reran each arm from the frozen task set. All 288 reported
responses parsed and reproduced the supplied evidence fields exactly. The raw
archive contains only these valid reported runs.

\section{Mechanical enforcement baseline}

Each experiment also records a non-LLM mechanical baseline. It substitutes the
known constrained optimum when a proposal violates the binding duty and otherwise
leaves the proposal unchanged. This is not a fair replacement for review in
settings where the optimum is unknown; it is a diagnostic upper bound for
constraint handling in our synthetic tasks. The evidence gate differs because
the reviewer must name a repair and accurately reproduce the certificate fields
before its recommendation gains force.

\section{Interpretation of the prevalence frontier}

For fixed conditional estimates, the zero contour in Figure~\ref{fig:frontier}
solves
\[
  \pi_C\delta_C+\pi_V\delta_V+(1-\pi_C-\pi_V)\delta_U=0.
\]
Points above or to the right of the contour favor the gate only when the
corresponding weighted sum is positive; geometric direction varies with the
signs of the estimated effects. The plot uses increments of 0.01 over the valid
simplex. It visualizes sensitivity to assumed base rates and does not attach
sampling uncertainty to the contour.

\end{document}